%% file: main.tex
\documentclass[10pt,twocolumn,letterpaper]{article}

\usepackage{iccv}
\usepackage{times}
\usepackage{epsfig}
\usepackage{graphicx}
\usepackage{amsmath}
\usepackage{amssymb}
\usepackage{float}
\usepackage{caption}

\input{macros}

\usepackage[pagebackref=true,breaklinks=true,letterpaper=true,colorlinks,bookmarks=false]{hyperref}

\iccvfinalcopy 

\def\iccvPaperID{2143} 

\ificcvfinal\fi
\begin{document}

\title{\OurName: Latent Cross-Consistency for Unpaired Image Translation}

\author{Omry Sendik\\
Tel Aviv University\\
Israel\\
{\tt\small}
\and
Dani Lischinski\\
Hebrew University Jerusalem\\
Israel\\
{\tt\small}
\and
Daniel Cohen-Or\\
Tel Aviv University\\
Israel\\
{\tt\small}
}

\twocolumn[{%
\renewcommand\twocolumn[1][]{#1}%
\maketitle
\begin{center}
    \centering
    \includegraphics[width=1\textwidth,height=3cm]{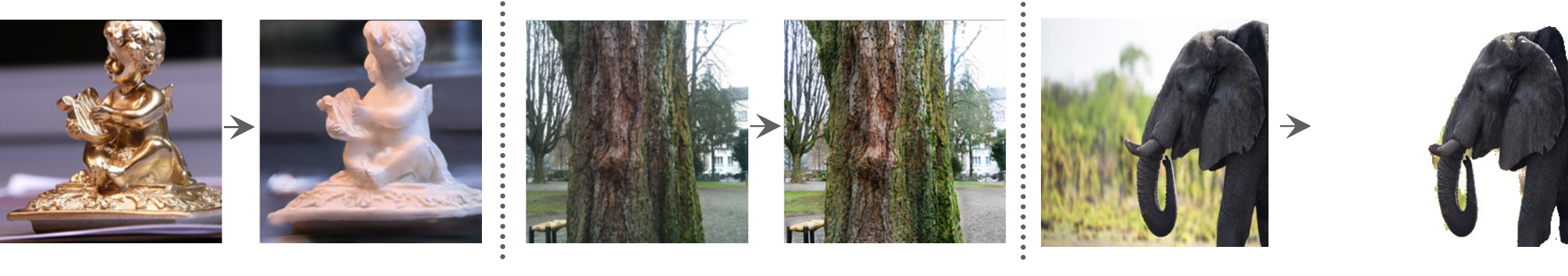}
    \captionof{figure}{Given two unpaired sets of images, we train a model to perform translation between the two sets. Here we show, from left to right, our results on changing a specular material to diffuse, enhancing a mobile phone image to look like one taken by a Digital SLR camera and foreground extraction.}
\end{center}%
\label{fig:teaser}
}]

\input{abstract}

\input{intro}
\input{priorart}

\input{overview}
\input{results}
\input{applications}
\input{appl_seg}
\input{conclusions}

{\small
\bibliographystyle{ieee}
\bibliography{bibliography}
}

\end{document}

%% file: macros.tex
\newcommand{\GANloss}{{\mathcal{L}_{\textit{GAN}}}}
\newcommand{\CTloss}{{\mathcal{L}_{\textit{CTC}}}}
\newcommand{\Idloss}{{\mathcal{L}_{\textit{Id}}}}
\newcommand{\ZIdloss}{{\mathcal{L}_{\textit{ZId}}}}
\newcommand{\ZCyloss}{{\mathcal{L}_{\textit{ZCyc}}}}
\newcommand{\lambdaGAN}{{\lambda_{\textit{GAN}}}}
\newcommand{\lambdaCT}{{\lambda_{\textit{CTC}}}}
\newcommand{\lambdaId}{{\lambda_{\textit{Id}}}}
\newcommand{\lambdaZId}{{\lambda_{\textit{ZId}}}}
\newcommand{\lambdaZCy}{{\lambda_{\textit{ZCyc}}}}

\newcommand{\Expect}[2]{\mathbb{E}_{x_{#1} \thicksim p_{\textit{data}}(#2)}}

\usepackage{xspace} 
\newcommand{\OurName}{CrossNet\xspace}
\newcommand{\FID}{$\phi$}
\newcommand{\BFFID}{$\pmb{\phi}$}

%% file: abstract.tex
\begin{abstract}

Recent GAN-based architectures have been able to deliver impressive performance on the general task of image-to-image translation. In particular, it was shown that a wide variety of image translation operators may be learned from two image sets, containing images from two different domains, without establishing an explicit pairing between the images. This was made possible by introducing clever regularizers to overcome the under-constrained nature of the unpaired translation problem.

In this work, we introduce a novel architecture for unpaired image translation, and explore several new regularizers enabled by it.
Specifically, our architecture comprises a pair of GANs, as well as a pair of translators between their respective latent spaces.
These \emph{cross-translators} enable us to impose several regularizing constraints on the learnt image translation operator, collectively referred to as \emph{latent cross-consistency}.
Our results show that our proposed architecture and latent cross-consistency constraints are able to outperform the existing state-of-the-art on a variety of image translation tasks.

\end{abstract}

%% file: intro.tex
\section{Introduction}
Many useful operations on images may be cast as an \emph{image translation} task.
These include style transfer, image colorization, automatic tone mapping  and many more. Several such operations are demonstrated in Figure~\ref{fig:teaser}1.
While each of these operations may be carried out by a carefully-designed task-specific operator, in many cases, the abundance of digital images along with the demonstrated effectiveness of deep learning architectures, makes a data-driven approach feasible and attractive. 

A straightforward supervised approach is to train a deep network to perform the task using a large number of pairs of images, before and after the translation \cite{isola2017image}.
However, collecting a large training set consisting of paired images is often prohibitively expensive or infeasible.

Alternatively, it has been demonstrated that an image translation operator, 
may also be learned from two image sets, containing images from two domains $A$ and $B$, respectively, without establishing an explicit pairing between images in the two sets \cite{zhu2017unpaired,yi2017dualgan,liu2016coupled}.
This is accomplished using generative adversarial networks (GANs) \cite{goodfellow2014generative}.

This latter approach is more attractive, as it requires much weaker supervision, however, this comes at the cost of making the translation problem highly under-constrained.
In particular, a meaningful pairing is not guaranteed, as there are many pairings that are able to yield the desired distribution of translated images.
Furthermore, undesirable phenomena, such as mode collapse, may arise when attempting to train the translation GAN \cite{goodfellow2014generative}.

To address these issues, existing GAN-based approaches for unpaired image translation \cite{zhu2017unpaired,yi2017dualgan}, train two GANs.
One GAN maps images from domain $A$ to domain $B$, and a second one operates in the opposite direction (from $B$ to $A$). 
Furthermore, a strong regularization is imposed in the form of the \emph{cycle consistency loss}, which ensures that concatenating the two translators roughly reconstructs the original image.
Note that the cycle consistency loss is measured using a pixelwise metric ($L_1$) in the original image domains.

\begin{figure}[t]
\centering
\includegraphics[width=1\linewidth]{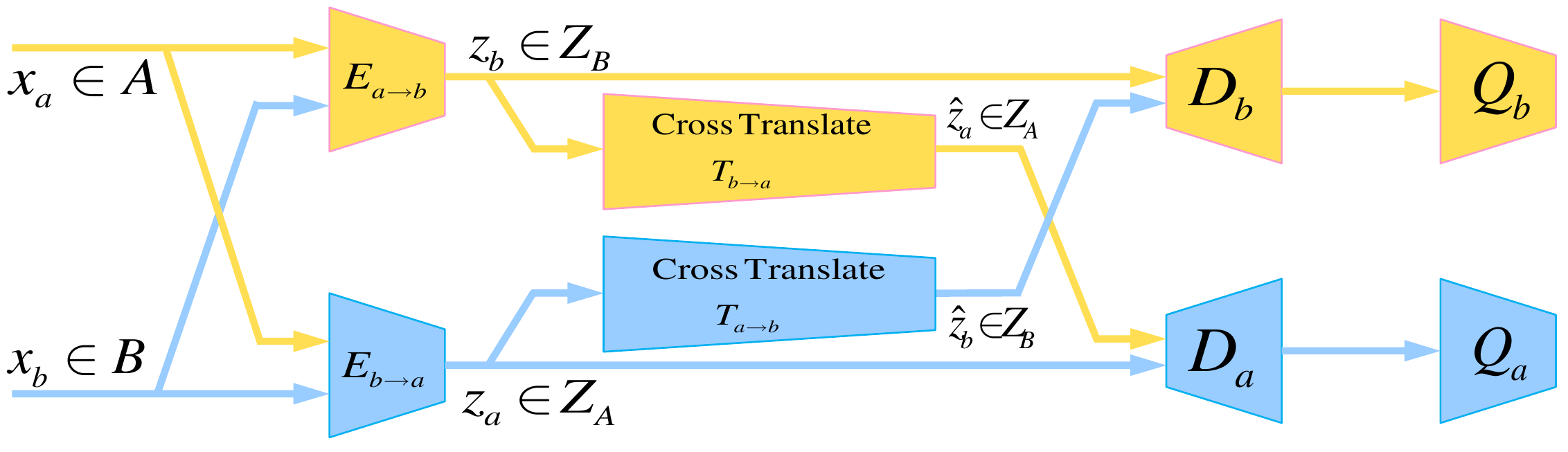}
\caption{Our architecture uses two GANs that learn an image translation operator from two unpaired sets of images, $A$ and $B$. By introducing a pair of \emph{cross-translators} between the latent spaces ($Z_B$ and $Z_A$) of the two Encoder-Decoder generators, we enable several novel latent cross-consistency constraints. Note that the only components used to perform the translation at test time are the encoders $E$ and the decoders $D$, while the other components (cross-translators $T$ and discriminators $Q$), as well as the crossing data paths, are only used by the training phase.
}
\label{fig:SymmArch}
\end{figure}

In this work, we introduce a novel architecture for unpaired image translation, and explore several new regularizers enabled by it.
Our architecture also comprises a pair of GANs (for $A \rightarrow B$ and $B \rightarrow A$ translation), but we also add a pair of translators between the latent spaces of their respective generators, $Z_B$ and $Z_A$, as shown in Figure~\ref{fig:SymmArch}.
These \emph{cross-translators} enable us to impose several regularizing constraints on the learnt image translation operator, collectively referred to as \emph{latent cross-consistency}.
Intuitively, regularizing the latent spaces is a powerful yet flexible approach: the latent representation computed by the GAN's generator captures the translation task's most pertinent information, while the original input representation (pixels) contains much additional irrelevant information.

We demonstrate the competence of the proposed architecture and latent cross-consistency, as well as several additional loss terms, through an ablation study and comparisons with existing approaches.
We show our competitive advantages vs.~the existing state-of-the-art on tasks such as translating between specular and diffuse objects, translating mobile phone photos to DSLR-like quality and foreground extraction. Figure \ref{fig:teaser}1 demonstrates some of our results.

%% file: priorart.tex
\section{Related work}
\subsection{Unpaired image-to-image translation}
2017 was a year with multiple breakthroughs in unpaired image-to-image translation.
Taigman et al.~\cite{taigman2016unsupervised} kicked off by proposing an unsupervised formulation employing GANs for transfer between two unpaired domains, demonstrating transfer of SVHN images to MNIST ones and of face photos from the Facescrub dataset to emojis.

Two seminal works which achieved great success in unpaired image-to-image translation are CycleGAN \cite{zhu2017unpaired} and DualGAN \cite{yi2017dualgan}. Both proposed to regularize the training procedure by requiring a bijection, enforcing the translation from source to target domain and back to the source to return to the same starting point. Such a constraint yields a meaningful mapping between the two domains. Furthermore, since bijection cannot be achieved in the case of mode collapse, it thus prevents it.

Dong et al.~\cite{dong2017unsupervised} trained a conditional GAN to learn shared global features from two image domains, then followed by synthesis of plausible images in either domain from a noise vector conditioned on a class/domain label. To enable image-to-image translation, they separately train an encoder to learn a mapping from an image to its latent code, which would serve as the noise input to the conditional GAN to generate a target image.

Choi et al.~\cite{choi2017stargan} proposed StarGAN, a network that learns the mappings among multiple domains using only a single generator and a discriminator. 
Kim et al.~\cite{kim2017learning} tackled the lack of image pairing in the image-to-image translation setting through a model based on two different GANs coupled together. Each GAN ensures that its generative function can map its domain to that of its counterpart. Since their method discovers relations between different domains, it may be leveraged to successfully transfer style.

A very different recent approach is NAM \cite{hoshen2018nam}, which relies on having a high quality pre-trained unsupervised generative model for the source domain. If such a generator is available, a generative model needs to be trained only once per target dataset, and can thus be used to map to many target domains without adversarial generative training.

In this work we also address unpaired image-to-image translation. Contemporary approaches and some of those mentioned above tackle this problem by imposing constraints formulated in the image domain. Our approach consists of novel regularizers operating \textit{across} the two latent spaces. Through the introduction of a unique architecture which enables a strong coupling between a pair of latent spaces, we are able to define a set of losses which are domain-agnostic.
Additionally, a benefit of our architecture is that it enables multiple regularizers, which together push the trained outcome to a more stable final result. We stress that this is different from contemporary approaches relying on image domain losses, which make use of one or two losses (usually an identity or cycle-consistency loss).
		
\subsection{Latent space regularization}

Motivated by the fact that image-to-image translation aims at learning a joint distribution of
images from the source and target domains, by using images from the marginal distributions in
individual domains, Liu et al.~\cite{liu2017unsupervised} made a shared-latent space assumption, and devised an architecture which maps images from both domains to the same latent space. By sharing weight parameters corresponding to high level semantics in both the encoder and decoder networks, the coupled GANs are enforced to interpret these image semantics in the same way.
Additionally, VeeGAN \cite{srivastava2017veegan} also addressed mode collapse by imposing latent space constraints. In their work, a reconstructor network reverses the action of the generator through an architecture which specifies a loss function over the latent space.

The two works mentioned above attempt to translate an image from a source domain $A$ to a single target domain $B$. The scheme by which they achieve this limits their ability to extend to translating an input image to multiple domains at once. Armed with this realization, Huang et al. \cite{huang2018multimodal} proposed a multimodal unpaired image-to-image translation (MUNIT) framework. Achieving this involved decoupling the latent space into content and style, under the assumption that what differs between target domains is the style alone.


Tackling the lack of diversity of the results, Lee et al.~\cite{lee2018diverse} proposed Diversifying Image-to-Image Translation (DRIT) by embedding images onto both a domain-invariant content space and a domain-specific attribute space.

Mejjati et al.~\cite{mejjati2018unsupervised} proposed AGGAN, improving image to image translation results by adding attention guidance, showing that their approach focuses on relevant regions in the image without requiring supervision.

In the entirety of these architectures there is no path within the network graph, which enables formulating losses that constrain both latent spaces at once. For this reason, we dissect the common GAN architecture, and propose a path between encoders and decoders from cross (opposite) domains. Our architecture thus consists of a pair of GANs, but in addition, we couple each generator with a translator between latent spaces. The addition of the translators opens up not only the ability to enforce bijection constraints in latent space but more intriguing losses, which further constrain the problem, leading to better translations.

%% file: overview.tex
\section{Cross Consistency Constraints}

\begin{figure}[ht]
\setlength\tabcolsep{1.5pt}
\begin{tabular}{cccc}

Input & Latent Code & Activations & Output\\

\includegraphics[width=0.24\linewidth]{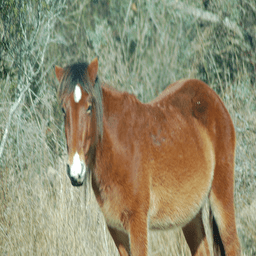} &  
\includegraphics[width=0.24\linewidth]{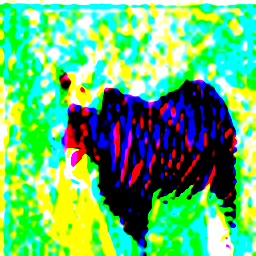} &
\includegraphics[width=0.24\linewidth]{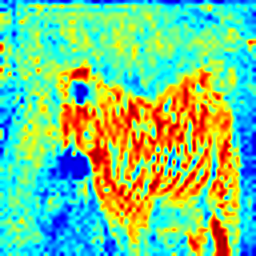} &
\includegraphics[width=0.24\linewidth]{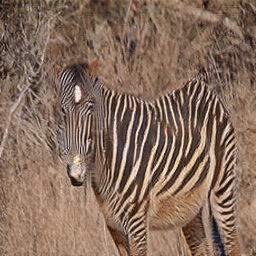} \\

\includegraphics[width=0.24\linewidth]{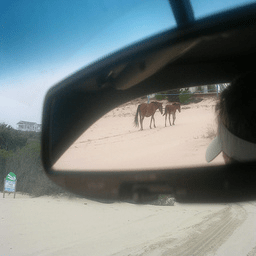} &  
\includegraphics[width=0.24\linewidth]{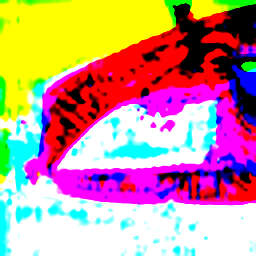} &
\includegraphics[width=0.24\linewidth]{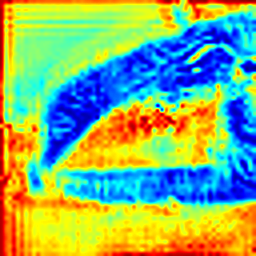} &
\includegraphics[width=0.24\linewidth]{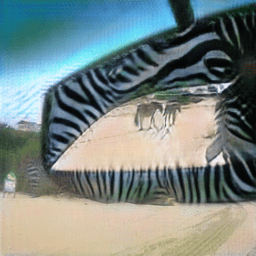} \\

\includegraphics[width=0.24\linewidth]{./Figures/PCA_n02381460_1090_real_A.png} &  
\includegraphics[width=0.24\linewidth]{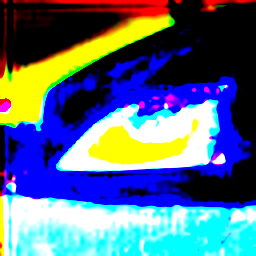} &
\includegraphics[width=0.24\linewidth]{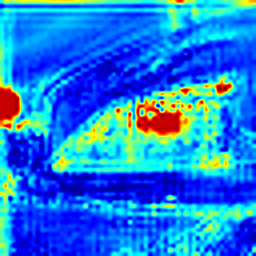} &
\includegraphics[width=0.24\linewidth]{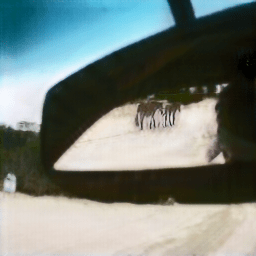} \\

\end{tabular}
\caption{
	Latent space visualization for two Horse-to-Zebra translation examples. The second column visualizes the latent space of the generator by using PCA to reduce the 256 channels of the latent space to three, mapped to RGB. Alternatively, the third column shows the magnitude of the 256-dimensional feature vector at each latent space neuron. Note that the latent space in these examples indicates the positions and shapes of the zebra stripes in the resulting translated image. The two upper rows show the latent space and results of CycleGAN. The bottom row shows the latent space and result in our approach, where the encoder does not attempt to texture improper regions, thanks to its training with our latent cross-consistency losses.
}
\label{tab:PCAMotiv}
\end{figure}

Architectures such as CycleGAN \cite{zhu2017unpaired} or DualGAN \cite{yi2017dualgan} are able to accomplish unpaired image-to-image translation by imposing consistency constraints in the original image domains $A$ and $B$.
Thus, their constraints operate on the original, pixel-based, image representations, which contain much information that is irrelevant to the translation task at hand.
However, it is well-known that in a properly trained Encoder-Decoder architecture, the latent space contains a distillation of the features that are the most relevant and pertinent to the task.
To demonstrate this, consider the Horse-to-Zebra translation task, for example. The top row of Figure \ref{tab:PCAMotiv} visualizes the latent code of CycleGAN's Horse-to-Zebra Encoder-Decoder generator, where we can see that the latent code already contains zebra-specific information, such as the locations and shapes of the zebra stripes.
In a manner of speaking, the generator's encoder has already planted ``all the makings of a zebra'' into the latent code, leaving the decoder with the relatively simpler task of transforming it back into the image domain.
Similarly, the second row of Figure~\ref{tab:PCAMotiv} demonstrates a case where the zebra-specific features are embedded in the wrong spatial regions, yielding a failed translation result.

The above observation motivates us to explore an architecture that enables imposing consistency constraints on and between the latent spaces. In our architecture, the latent spaces of the two Encoder-Decoder generators (from $A$ to $B$ and vice versa) are coupled via a pair of \emph{cross-translators}. This creates additional paths through which data can flow during training, enabling several novel latent cross-consistency constraints in the training stage, as described below. The bottom row of Figure~\ref{tab:PCAMotiv} shows an example where our encoder, trained with these constraints, avoids the incorrect embedding of the zebra specific features.

\subsection{Architecture}
Armed with the motivation to impose regularizations in latent space, we propose an architecture which links between the latent spaces of the generators (from domain $A$ to $B$ and vice versa), thus enabling a variety of latent consistency constraints.
Our architecture is shown in Figure~\ref{fig:SymmArch}.

The architecture consists of a pair of Encoder-Decoder generators, which we denote by $(E_{a\rightarrow b},D_{b})$, and $(E_{b\rightarrow a},D_{a})$.
The encoder $E_{a\rightarrow b}$ encodes an image $x_{a} \in A$ to a latent code denoted by $z_b \in Z_B$, while $D_{b}$ decodes it to an output image $\hat{x}_{b}$. Similarly, the encoder $E_{b\rightarrow a}$ encodes an image $x_{b} \in B$ to a latent code denoted by $z_a \in Z_A$, while $D_{a}$ decodes it to an output image $\hat{x}_{a}$.
The discriminators, $Q_{b}$ and $Q_{a}$, attempt to determine whether or not an input image from domains $B$ or $A$, respectively, is real or fake. 
The novel part of our architecture is the addition of two \emph{cross-translators}, $T_{b \rightarrow a}$ and $T_{a \rightarrow b}$, shown in Figure \ref{fig:SymmArch}.
Each translator is trained to transform the latent codes of one generator into those of the other, namely, from $z_b$ to $\hat{z}_{a}$ and from $z_a$ to $\hat{z}_{b}$, respectively.
By adding the two cross-translators to our architecture, several additional paths, through which data may flow during training, become possible, paving the way for new consistency constraints.
In this work, we present three novel latent cross-consistency losses, which are shown to conjoin to produce superior results on a variety of image translation tasks.

Note that the cross-translators $T$ and the discriminators $Q$ are only used at train time. At test time, the only components used for translating images are the encoders $E$ and the decoders $D$.

%

\subsection{Latent Cross-Identity Loss}

In order to train the cross-translators $T_{a\rightarrow b}$ and $T_{b\rightarrow a}$, we require that an image $x_a$ fed into the encoder $E_{a\rightarrow b}$ should be reconstructable by the decoder of the dual generator, $D_a$, after translation of its latent code by $T_{b\rightarrow a}$. A symmetric requirement is imposed on the translator $T_{a\rightarrow b}$. These two requirements are formulated as the \emph{latent cross-identity loss}:
\begin{equation}
\label{eq:zidloss}
\begin{split}
\ZIdloss = \Expect{a}{A}\lbrack\Vert{D_a(T_{b \rightarrow a}(E_{a \rightarrow b}(x_a)))-x_a}\Vert{_1}\rbrack + \\
\Expect{b}{B}\lbrack\Vert{D_b(T_{a \rightarrow b}(E_{b \rightarrow a}(x_b)))-x_b}\Vert{_1}\rbrack.
\end{split}
\end{equation}
The corresponding data path through the network is shown in Figure \ref{fig:Archs}(a).
This may be thought of as an autoencoder loss, where the autoencoder, in addition to an encoder and a decoder, has a bipartite latent space, with a translator between its two parts. 

Note that some previous unpaired translation works \cite{yi2017dualgan,zhu2017unpaired}, use an ordinary identity loss (without cross-translation), where images from domain $B$ are fed to the $A \rightarrow B$ generator, and vice versa. We adopt this loss as well, as we found it to complement our cross-identity loss in Eq.~\eqref{eq:zidloss}: 
\begin{equation}
\label{eq:idloss}
\begin{split}
\Idloss = \Expect{a}{A}\lbrack\Vert{D_a(E_{b \rightarrow a}(x_a))-x_a}\Vert{_1}\rbrack + \\
\Expect{b}{B}\lbrack\Vert{D_b(E_{a \rightarrow b}(x_b))-x_b}\Vert{_1}\rbrack.
\end{split}
\end{equation}

\subsection{Latent Cross-Translation Consistency}

\begin{figure}[ht]
\centering
\includegraphics[width=0.85\linewidth]{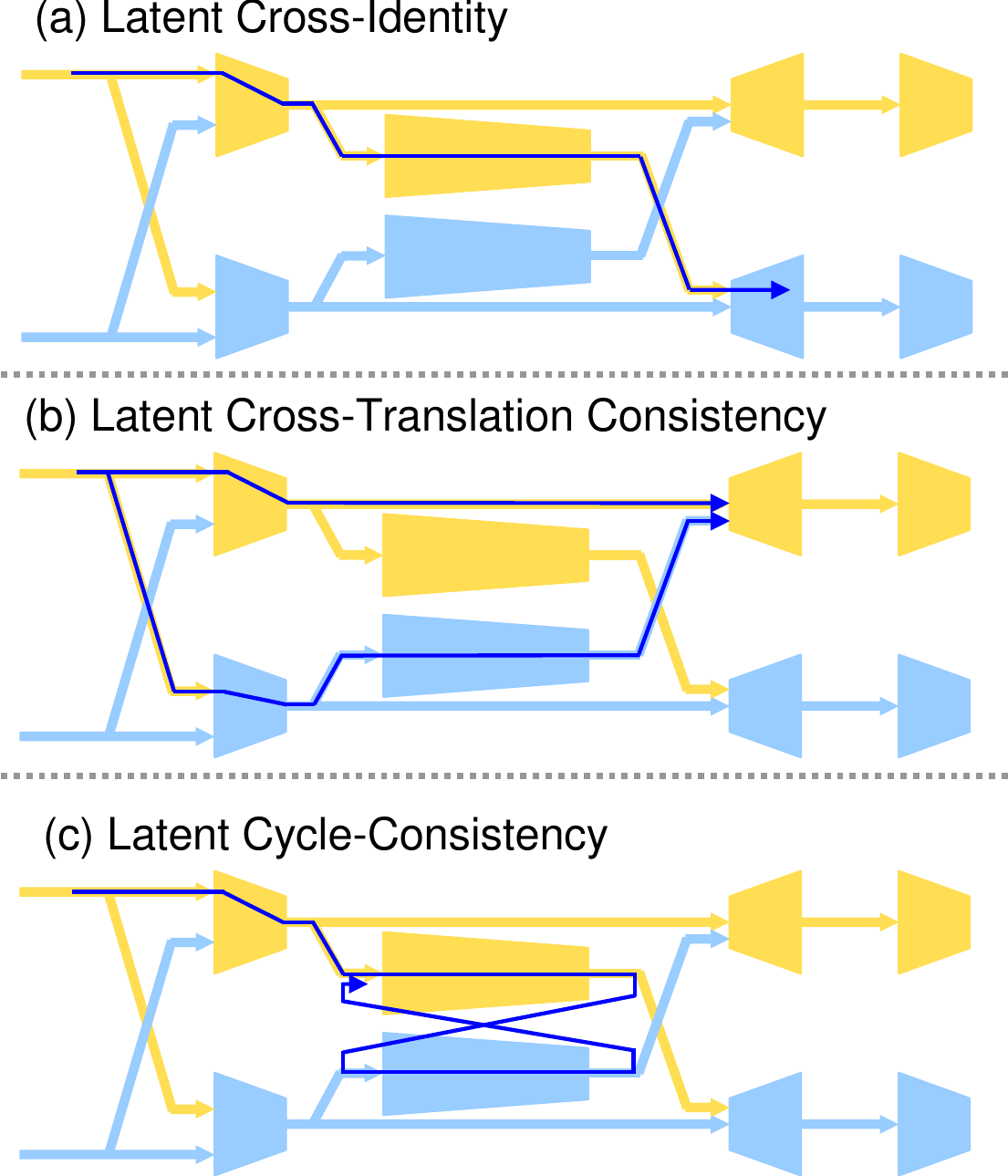}
\caption{Data paths used by the novel loss terms when training our model (symmetric paths are omitted for clarity). (a) The latent cross-identity loss trains the cross-translators to map between the two latent spaces of the dual generators.
(b) The latent cross-translation consistency loss regularizes the latent spaces generated by each of the two encoders.
(c) The latent cycle-consistency loss ensures that the cross-translators define bijections between the two latent spaces.}
\label{fig:Archs}
\end{figure}

While the normal expected input, for each encoder, is an image from its intended source domain, let us consider the scenario where one of the encoders is given an image from its target domain, instead.
For example, if $A$ are images of horses and $B$ images of zebras, what should happen when a zebra image $x_b$ is given as input to the ``horse-to-zebra'' encoder $E_{a\rightarrow b}$?
Our intuition tells us that in such a case we'd like the generator to avoid modifying its input.
This implies that the resulting latent code $z_b = E_{a\rightarrow b}(x_b)$ should capture and retain the essential ``zebra-specific'' information present in the input image.
The translator $T_{b\rightarrow a}$ is trained to map such ``zebra features'' to ``horse features'', thus we expect $\hat{z}_a = T_{b\rightarrow a}(z_b)$ to be similar to the latent code $z_a = E_{b\rightarrow a}(x_b)$, obtained by feeding the zebra image to the ``zebra-to-horse'' encoder, which should also yield ``horse features''.
The above reasoning, applied in both directions, is formally expressed using the \emph{latent cross-translation consistency loss} (see Figure \ref{fig:Archs}(b)):


\begin{equation}
\begin{split}
\CTloss = \Expect{a}{A}\lbrack\Vert{T_{a \rightarrow b}(E_{b \rightarrow a}(x_{a}))-E_{a \rightarrow b}(x_{a})}\Vert{_1}\rbrack + \\
\Expect{b}{B}\lbrack\Vert{T_{b \rightarrow a}(E_{a \rightarrow b}(x_{b}))-E_{b \rightarrow a}(x_{b})}\Vert{_1}\rbrack.
\end{split}
\end{equation}


\vspace{-4.25mm}
\subsection{Latent Cycle-Consistency}

Our final latent space regularization is designed to ensure that our cross-translators are bijections between the two latent spaces of the generators. Similarly to the motivation behind the cycle-consistency loss of Zhu et al.~\cite{zhu2017unpaired}, having bijections helps achieving a meaningful mapping between the two domains, as well as avoids mode collapse during the optimization process.

Specifically, we require that translating a latent code $z_b$ first by $T_{b \rightarrow a}$ and then back by $T_{a \rightarrow b}$ yields roughly the same code back:
\begin{equation}
\begin{split}
\ZCyloss = \Expect{a}{A}\lbrack\Vert{T_{a \rightarrow b}(T_{b \rightarrow a}(z_b))-z_b}\Vert{_1}\rbrack+ \\
\Expect{b}{B}\lbrack\Vert{T_{b \rightarrow a}(T_{a \rightarrow b}(z_a))-z_a}\Vert{_1}\rbrack.
\end{split}
\end{equation}
The data path corresponding to this \emph{latent cycle-consistency loss} is depicted in Figure \ref{fig:Archs}(c).

\subsection{Implementation and Training}
\label{sec:TrainDetails}
Our final loss, which we optimize throughout the entire training process is a weighed sum of the losses presented in the previous sections, and a GAN loss \cite{goodfellow2014generative},

\begin{equation}
\label{eq:full-loss}
\begin{split}
\mathcal{L} & = \lambdaGAN \GANloss + \lambdaId \Idloss +\\
& \lambdaCT \CTloss + \lambdaZId \ZIdloss + \lambdaZCy \ZCyloss.
\end{split}
\end{equation}

Rather than using a negative log likelihood objective in the GAN loss, we make use of a least-squares loss \cite{mao2017least}.
Additionally, we adopt Shrivastava et al.'s method ~\cite{shrivastava2017learning}, which updates the discriminators using a history of translated images rather than only the recently translated ones.

Unless otherwise mentioned, in all of the experiments in this paper, we set $\lambdaGAN=1$, $\lambdaId=3$, $\lambdaCT=3$, $\lambdaZId=6$ and $\lambdaZCy=6$.

For our generators, we adopt the Encoder and Decoder architecture from Johnson et al.~\cite{johnson2016perceptual}. Their Encoder architecture consists of an initial 7x7 convolution with stride 1, two stride 2 convolutions with a 3x3 kernel, and 9 residual blocks with 3x3 convolutions. The Decoder consists of two transposed convolutions with stride 2 and a kernel size of 3x3, followed by a final convolution with kernel size of 7x7 and hyperbolic tangent activations for normalization of the output range. For our discriminator networks, we use 70x70 PatchGANs \cite{li2016precomputed}, whose task is to classify whether the overlapping patches are fake or real.

Finally, our two latent code translators consist of 9 residual blocks that use 3x3 convolutions with stride 1.

For all of the applications which we present in the following sections, we trained our proposed method using two sets of \textasciitilde 1000 images (a total of \textasciitilde 2000 images). The final generator which we used for producing the results was selected as the result after training for 200 epochs.

%% file: results.tex
\section{Comparisons}
\subsection{Ablation study}

\begin{figure}[ht]
\setlength\tabcolsep{0.5pt}
\begin{tabular}{ccccccc}

Input & $\CTloss$ & $\ZIdloss$ & $\small\begin{array} {lcl} \CTloss \\ \ZIdloss \end{array}$ & $\small\begin{array} {lcl} \ZIdloss \\ \ZCyloss \end{array}$ &
$\small\begin{array} {lcl} \CTloss \\ \ZIdloss \\ \ZCyloss \end{array}$ &
\footnotesize{CycGAN} \\

\multicolumn{7}{c}{Zebra to Horse} \\

\raisebox{-.75\height}{\includegraphics[width=0.13\linewidth]{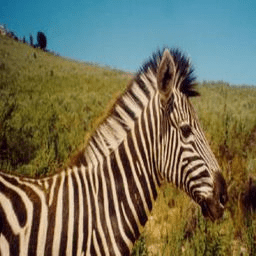}} &
\raisebox{-.75\height}{\includegraphics[width=0.13\linewidth]{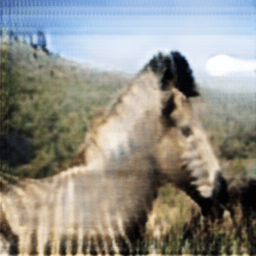}}  &
\raisebox{-.75\height}{\includegraphics[width=0.13\linewidth]{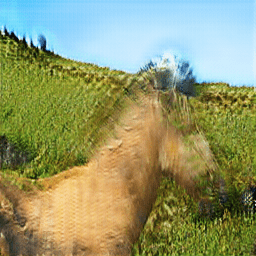}} &
\raisebox{-.75\height}{\includegraphics[width=0.13\linewidth]{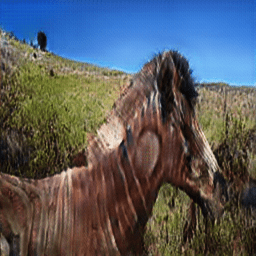}} &
\raisebox{-.75\height}{\includegraphics[width=0.13\linewidth]{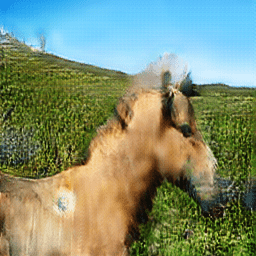}} &
\raisebox{-.75\height}{\includegraphics[width=0.13\linewidth]{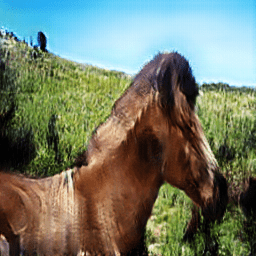}} &
\raisebox{-.75\height}{\includegraphics[width=0.13\linewidth]{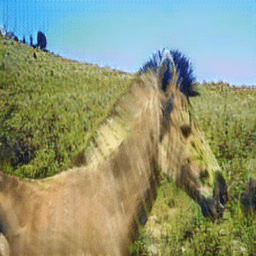}} \\

\raisebox{-.75\height}{\includegraphics[width=0.13\linewidth]{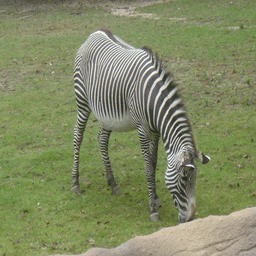}} &
\raisebox{-.75\height}{\includegraphics[width=0.13\linewidth]{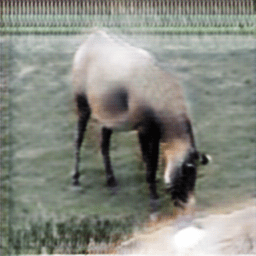}} &
\raisebox{-.75\height}{\includegraphics[width=0.13\linewidth]{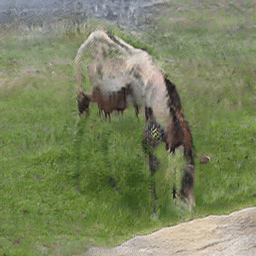}} &
\raisebox{-.75\height}{\includegraphics[width=0.13\linewidth]{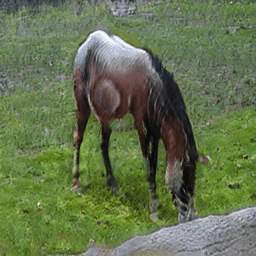}} &
\raisebox{-.75\height}{\includegraphics[width=0.13\linewidth]{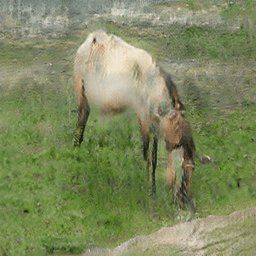}} &
\raisebox{-.75\height}{\includegraphics[width=0.13\linewidth]{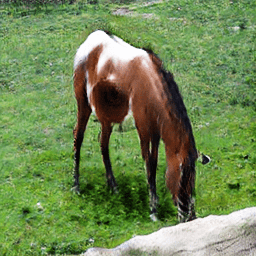}} &
\raisebox{-.75\height}{\includegraphics[width=0.13\linewidth]{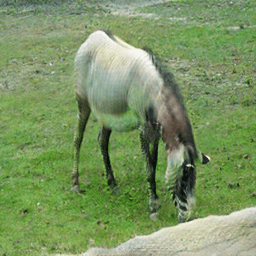}} \\

\raisebox{-.75\height}{\includegraphics[width=0.13\linewidth]{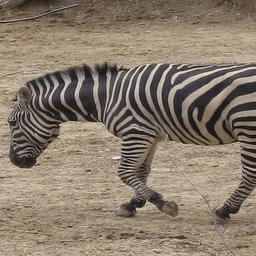}} &
\raisebox{-.75\height}{\includegraphics[width=0.13\linewidth]{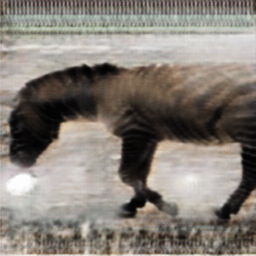}} & 
\raisebox{-.75\height}{\includegraphics[width=0.13\linewidth]{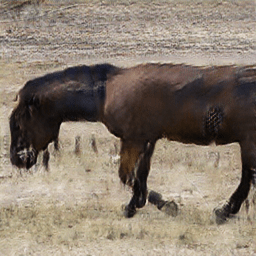}} & 
\raisebox{-.75\height}{\includegraphics[width=0.13\linewidth]{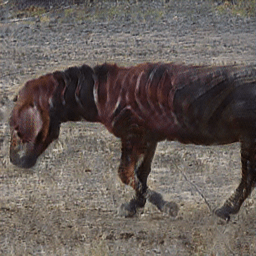}} & 
\raisebox{-.75\height}{\includegraphics[width=0.13\linewidth]{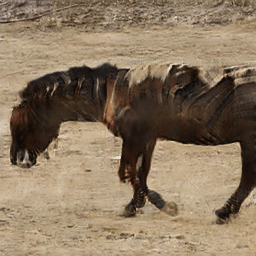}} & 
\raisebox{-.75\height}{\includegraphics[width=0.13\linewidth]{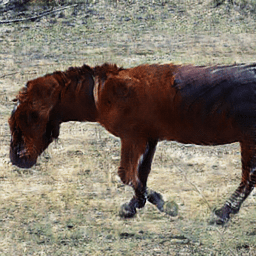}} & 
\raisebox{-.75\height}{\includegraphics[width=0.13\linewidth]{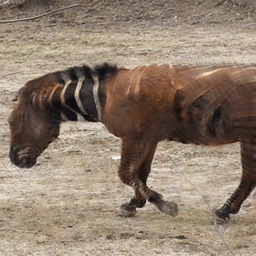}} \\

& \footnotesize{\FID=89.3} & \footnotesize{\FID=88.7} & \footnotesize{\FID=85.9} & \footnotesize{\FID=84.8} & \footnotesize{\textbf{\BFFID=81.2}} & \footnotesize{\FID=142.2} \\
 
\multicolumn{7}{c}{Horse to Zebra} \\

\raisebox{-.75\height}{\includegraphics[width=0.13\linewidth]{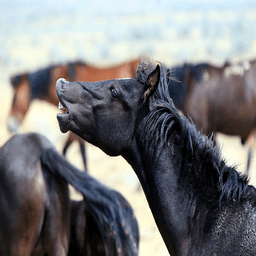}} &
\raisebox{-.75\height}{\includegraphics[width=0.13\linewidth]{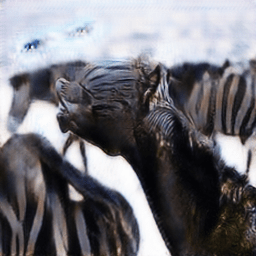}} &
\raisebox{-.75\height}{\includegraphics[width=0.13\linewidth]{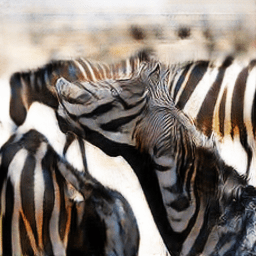}} &
\raisebox{-.75\height}{\includegraphics[width=0.13\linewidth]{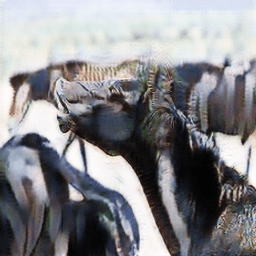}} &
\raisebox{-.75\height}{\includegraphics[width=0.13\linewidth]{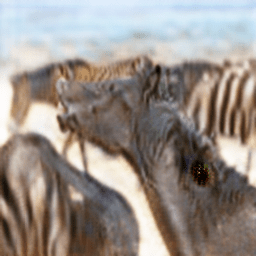}} &
\raisebox{-.75\height}{\includegraphics[width=0.13\linewidth]{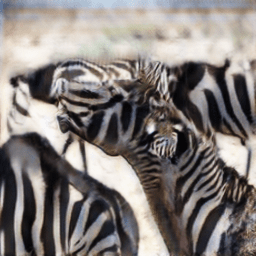}} &
\raisebox{-.75\height}{\includegraphics[width=0.13\linewidth]{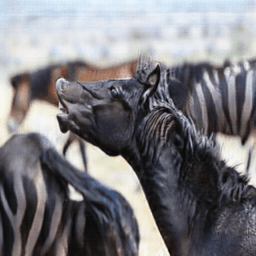}} \\

\raisebox{-.75\height}{\includegraphics[width=0.13\linewidth]{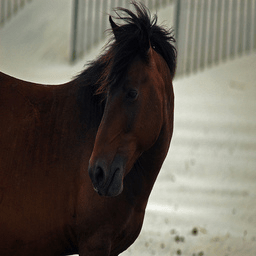}} & 
\raisebox{-.75\height}{\includegraphics[width=0.13\linewidth]{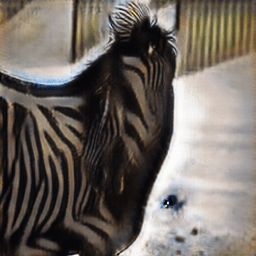}} &
\raisebox{-.75\height}{\includegraphics[width=0.13\linewidth]{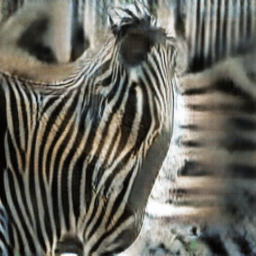}} &
\raisebox{-.75\height}{\includegraphics[width=0.13\linewidth]{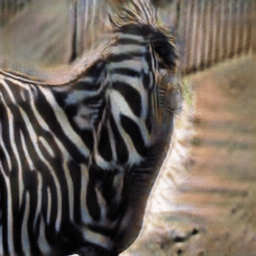}} &
\raisebox{-.75\height}{\includegraphics[width=0.13\linewidth]{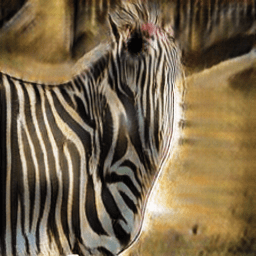}} & 
\raisebox{-.75\height}{\includegraphics[width=0.13\linewidth]{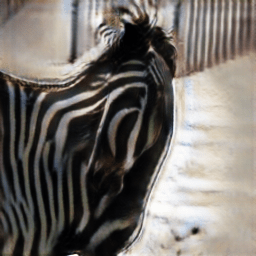}} &
\raisebox{-.75\height}{\includegraphics[width=0.13\linewidth]{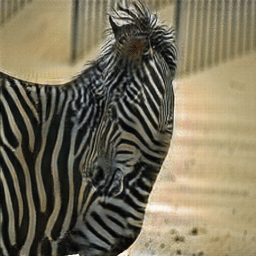}} \\

\raisebox{-.75\height}{\includegraphics[width=0.13\linewidth]{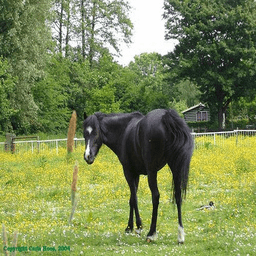}} &
\raisebox{-.75\height}{\includegraphics[width=0.13\linewidth]{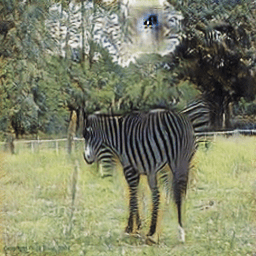}} &
\raisebox{-.75\height}{\includegraphics[width=0.13\linewidth]{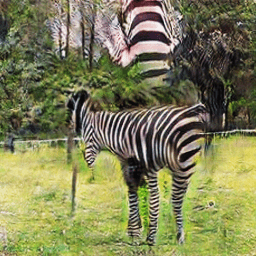}} &
\raisebox{-.75\height}{\includegraphics[width=0.13\linewidth]{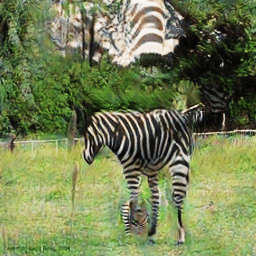}} &
\raisebox{-.75\height}{\includegraphics[width=0.13\linewidth]{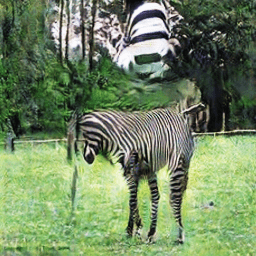}} &
\raisebox{-.75\height}{\includegraphics[width=0.13\linewidth]{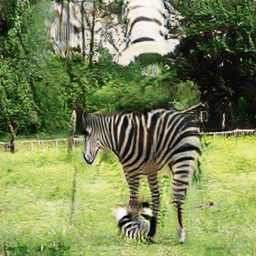}} &
\raisebox{-.75\height}{\includegraphics[width=0.13\linewidth]{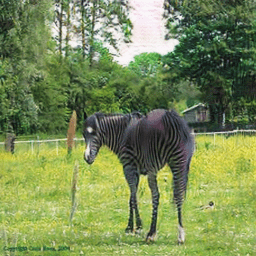}} \\

& \footnotesize{\FID=267.8} & \footnotesize{\FID=185.2} & \footnotesize{\FID=179.5} & \footnotesize{\FID=181.1} & \footnotesize{\textbf{\BFFID=154.4}} & \footnotesize{\FID=173.1} \\

\end{tabular}

\caption{Ablation study: Through a gradual inclusion of losses, we demonstrate how the results of various translation tasks improve. A combination of all three cross-consistency losses (6th column) is shown to yield better results than those produced by CycleGAN (7th column) for the Horse $\leftrightarrow$ Zebra translations.}
\label{tab:AblStdy}
\end{figure}

For evaluating the competence and effect of our newly devised cross-consistency losses, we conduct an ablation study. In Figure \ref{tab:AblStdy} we show some results for the Horse $\leftrightarrow$ Zebra translation, demonstrating the visual effect of adding each one of our three latent space losses. Additionally, we compare our results to those of CycleGAN \cite{zhu2017unpaired}. In all of these results, both $\GANloss$ and $\Idloss$ were included in our training.
We use the Frechet Inception Distance (FID) \cite{heusel2017gans} as an objective quantitative measure of result quality. We calculate the FID between a synthesized set and a set of real images from the target domain.

As may be seen in Figure \ref{tab:AblStdy}, through gradually adding the three cross-consistency losses, results improve. The best results are obtained when all three losses are included, as shown in the 6th column. The 7th (rightmost) column shows translation results generated using CycleGAN, where it may be seen that they are less successful: some zebra stripes remain when translating zebras to horses (3rd row), and not all horses are translated to zebras (rows 4,6). The FIDs decrease as we add more of our loss terms, reaching our best result once all 3 losses conjoin. It may be seen that the FIDs of our results are better that those of CycleGAN.

\subsection{Unpaired image-to-image translation}

\begin{figure*}[ht]
\begin{tabular}{lccc}
& \hspace{5px} Input \hspace{10px} \OurName  \hspace{10px} CycGAN & Input \hspace{10px} \OurName  \hspace{10px} CycGAN & Input \hspace{10px} \OurName  \hspace{10px} CycGAN \\

\vtop{\hbox{\strut Apple to Orange}\hbox{\OurName \FID=218.21}\hbox{CycGAN \FID=283.18}} &
\raisebox{-.75\height}{\includegraphics[width=0.09\linewidth]{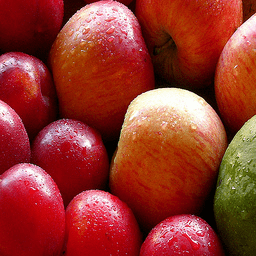}} \raisebox{-.75\height}{\includegraphics[width=0.09\linewidth]{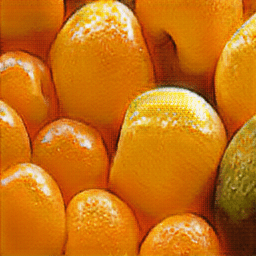}}
\raisebox{-.75\height}{\includegraphics[width=0.09\linewidth]{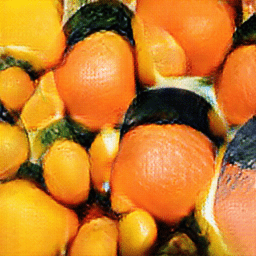}} & 
\raisebox{-.75\height}{\includegraphics[width=0.09\linewidth]{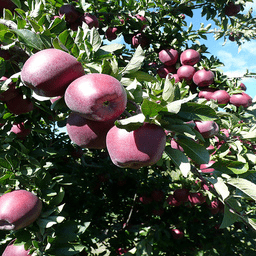}} \raisebox{-.75\height}{\includegraphics[width=0.09\linewidth]{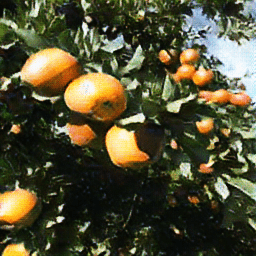}}
\raisebox{-.75\height}{\includegraphics[width=0.09\linewidth]{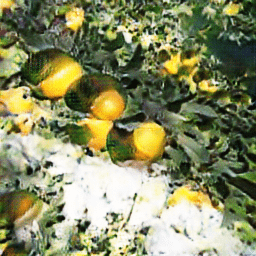}} &
\raisebox{-.75\height}{\includegraphics[width=0.09\linewidth]{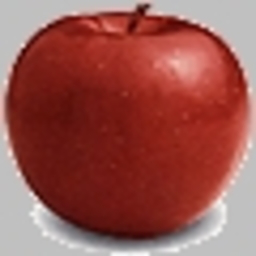}} \raisebox{-.75\height}{\includegraphics[width=0.09\linewidth]{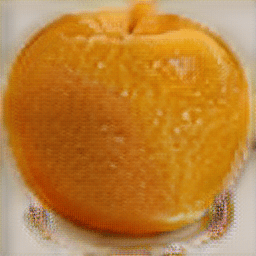}}
\raisebox{-.75\height}{\includegraphics[width=0.09\linewidth]{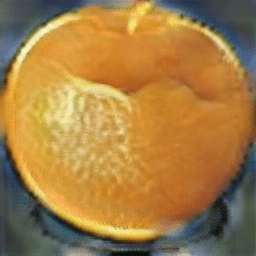}} \\

\vtop{\hbox{\strut Orange to Apple}\hbox{\OurName \FID=213.51}\hbox{CycGAN \FID=265.53}} &
\raisebox{-.75\height}{\includegraphics[width=0.09\linewidth]{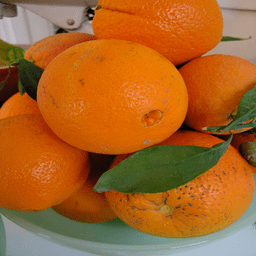}} \raisebox{-.75\height}{\includegraphics[width=0.09\linewidth]{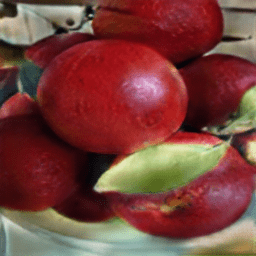}}
\raisebox{-.75\height}{\includegraphics[width=0.09\linewidth]{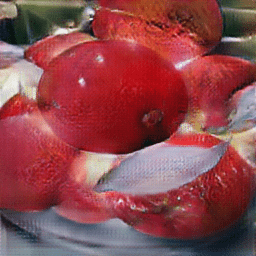}} &
\raisebox{-.75\height}{\includegraphics[width=0.09\linewidth]{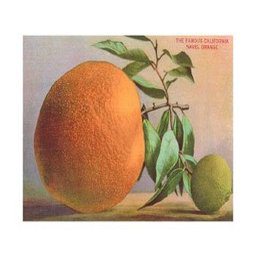}} \raisebox{-.75\height}{\includegraphics[width=0.09\linewidth]{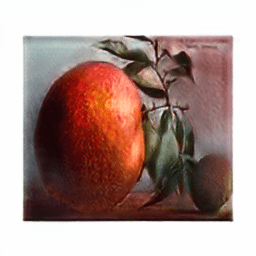}}
\raisebox{-.75\height}{\includegraphics[width=0.09\linewidth]{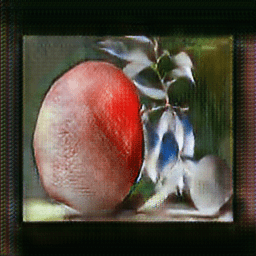}} &
\raisebox{-.75\height}{\includegraphics[width=0.09\linewidth]{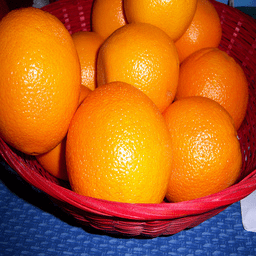}} \raisebox{-.75\height}{\includegraphics[width=0.09\linewidth]{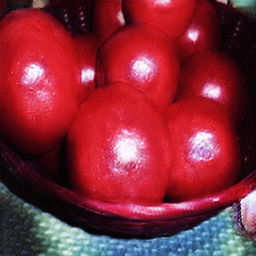}}
\raisebox{-.75\height}{\includegraphics[width=0.09\linewidth]{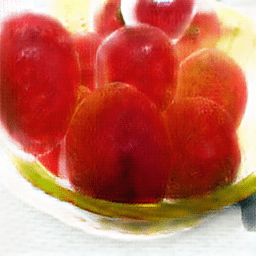}} \\

\vtop{\hbox{\strut Summer to Winter}\hbox{\OurName \FID=217.73}\hbox{CycGAN \FID=219.83}} &
\raisebox{-.75\height}{\includegraphics[width=0.09\linewidth]{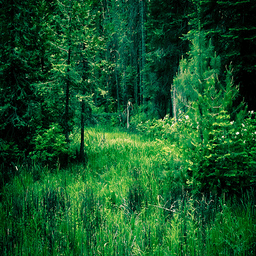}} \raisebox{-.75\height}{\includegraphics[width=0.09\linewidth]{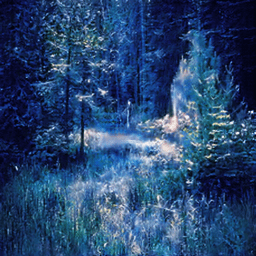}}
\raisebox{-.75\height}{\includegraphics[width=0.09\linewidth]{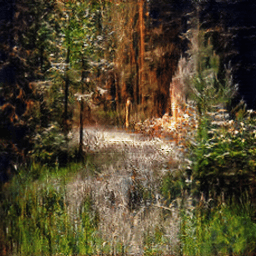}} &
\raisebox{-.75\height}{\includegraphics[width=0.09\linewidth]{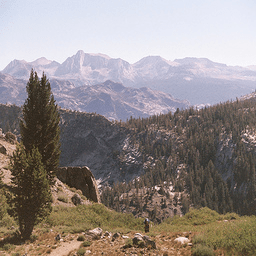}} \raisebox{-.75\height}{\includegraphics[width=0.09\linewidth]{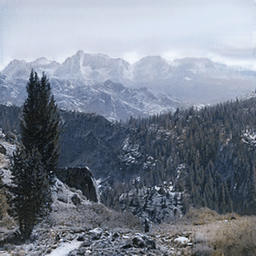}}
\raisebox{-.75\height}{\includegraphics[width=0.09\linewidth]{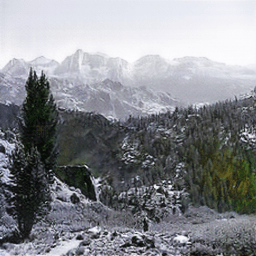}} &
\raisebox{-.75\height}{\includegraphics[width=0.09\linewidth]{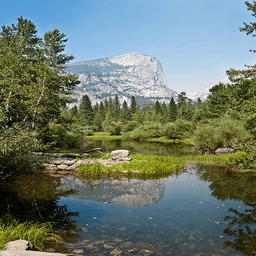}} \raisebox{-.75\height}{\includegraphics[width=0.09\linewidth]{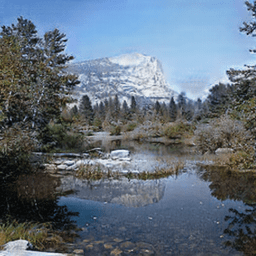}}
\raisebox{-.75\height}{\includegraphics[width=0.09\linewidth]{./Figures/2011-08-11_153210_fake_B_IDT3_T3_TIDT6_TCYC6.png}}  \\

\vtop{\hbox{\strut Winter to Summer}\hbox{\OurName \FID=130.40}\hbox{CycGAN \FID=137.51}} &
\raisebox{-.75\height}{\includegraphics[width=0.09\linewidth]{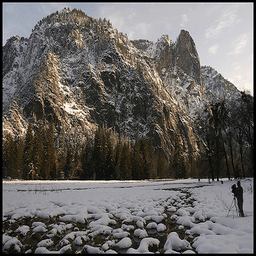}} \raisebox{-.75\height}{\includegraphics[width=0.09\linewidth]{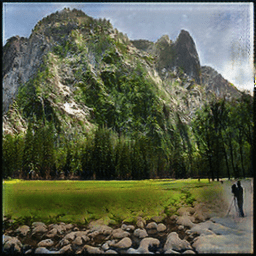}}
\raisebox{-.75\height}{\includegraphics[width=0.09\linewidth]{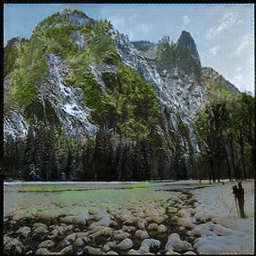}} &
\raisebox{-.75\height}{\includegraphics[width=0.09\linewidth]{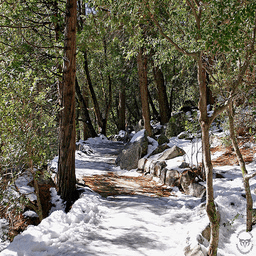}} \raisebox{-.75\height}{\includegraphics[width=0.09\linewidth]{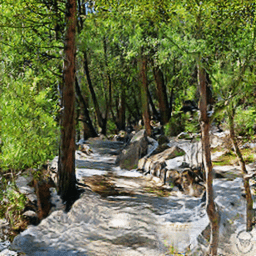}}
\raisebox{-.75\height}{\includegraphics[width=0.09\linewidth]{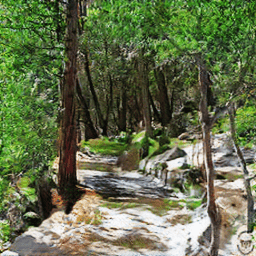}} &
\raisebox{-.75\height}{\includegraphics[width=0.09\linewidth]{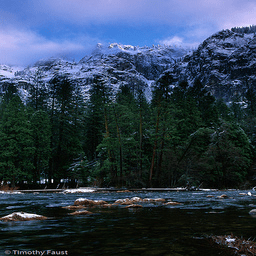}} \raisebox{-.75\height}{\includegraphics[width=0.09\linewidth]{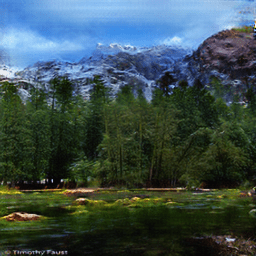}}
\raisebox{-.75\height}{\includegraphics[width=0.09\linewidth]{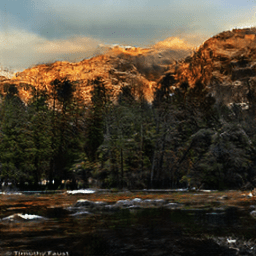}} \\

\end{tabular}
\caption{A variety of image-to-image translation results of our method (\OurName), compared to CycleGAN. In the Apples $\rightarrow$ Oranges translations CycleGAN consistently produces errors and artifacts that seem semantic in nature (shadows become illuminated, the rightmost orange looks like a superimposed image of an open orange).}
\label{tab:c2VsCG}
\end{figure*}

In Figure \ref{tab:c2VsCG} we show a variety of our unpaired image-to-image translation results (\OurName), compared with CycleGAN. From top to bottom, we show image-to-image translation results for: Apples $\leftrightarrow$ Oranges and Summer $\leftrightarrow$ Winter images. All of our results were achieved with the full loss in \eqref{eq:full-loss}, with the relative weights reported in Section~\ref{sec:TrainDetails}. Qualitatively, it may be observed that in all three image-to-image translation tasks, \OurName outperforms CycleGAN, providing better texture transfer, color reproduction, and also better structure (visible in the Apples $\leftrightarrow$ Oranges translations). Additionally, the FID shows superior results for \OurName. Note that for producing the CycleGAN results, where possible, we used the existing pretrained models, made available by Zhu et al.~\cite{zhu2017unpaired}.

\setlength\tabcolsep{0.6pt}


%% file: applications.tex
\section{Applications}

\subsection{Specular $\leftrightarrow$ Diffuse}
Most multi-view 3D reconstruction algorithms, assume that object appearance is predominantly diffuse. However, in real world images, often the contrary is true. In order to alleviate this restriction, Wu et al.~\cite{wu2018specular} propose a neural network for transferring multiple views of objects with specular reflection into diffuse ones. By introducing a task specific loss, exploiting the multiple views of a single specular object, they are able to synthesize faithful diffuse appearances of an object.
We make use of their publicly available dataset, and train \OurName to translate between specular and diffuse objects. 

Please note that differently from Wu et al., we do not rely on any prior assumptions specific to the task of specular to diffuse translation. A fundamental difference between Wu et al. and our approach is that the input to their network is an image sequence, which encourages the learning of a specific object's structure, while in our approach, only a single image is provided as input to the appropriate generator.




\setlength\tabcolsep{0.5pt}
\begin{figure}[th]
\begin{tabular}{ccccc}

\multicolumn{5}{c}{Specular to Diffuse} \\
Input & \OurName & CycleGAN & DRIT & MUNIT \\

\includegraphics[width=0.19\linewidth]{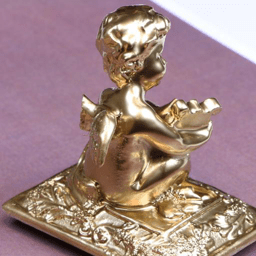} &
\includegraphics[width=0.19\linewidth]{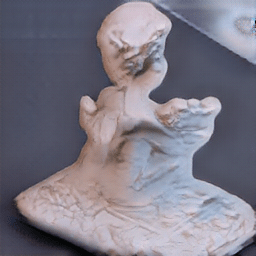} &
\includegraphics[width=0.19\linewidth]{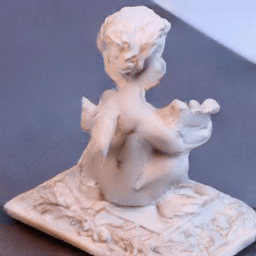} &
\includegraphics[width=0.19\linewidth]{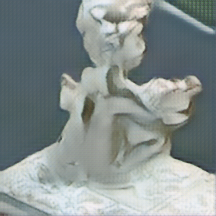} &
\includegraphics[width=0.19\linewidth]{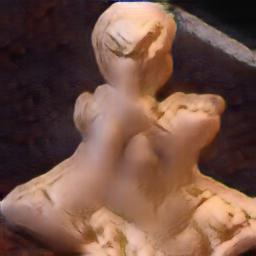} \\

\includegraphics[width=0.19\linewidth]{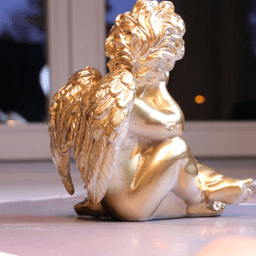} &
\includegraphics[width=0.19\linewidth]{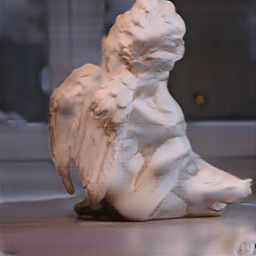} &
\includegraphics[width=0.19\linewidth]{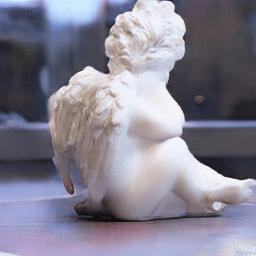} &
\includegraphics[width=0.19\linewidth]{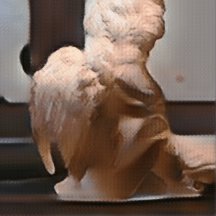} &
\includegraphics[width=0.19\linewidth]{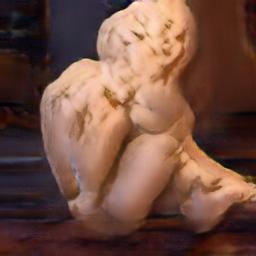} \\

& \FID=\textbf{239.17} & \FID=264.93 & \FID=284.42 & \FID=316.12\\

\multicolumn{5}{c}{Diffuse to Specular} \\

\includegraphics[width=0.19\linewidth]{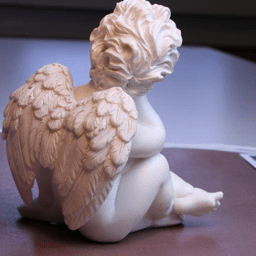} &
\includegraphics[width=0.19\linewidth]{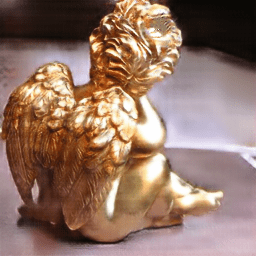} &
\includegraphics[width=0.19\linewidth]{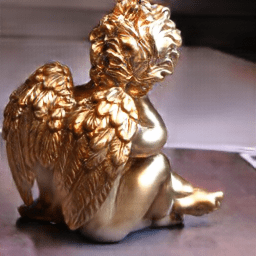} &
\includegraphics[width=0.19\linewidth]{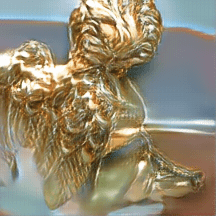} &
\includegraphics[width=0.19\linewidth]{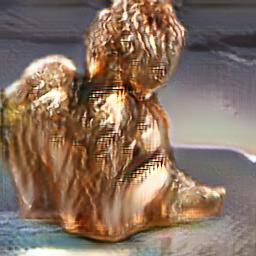} \\

\includegraphics[width=0.19\linewidth]{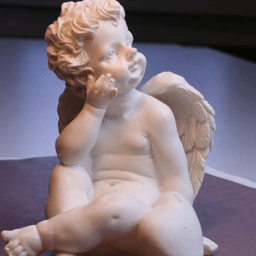} &
\includegraphics[width=0.19\linewidth]{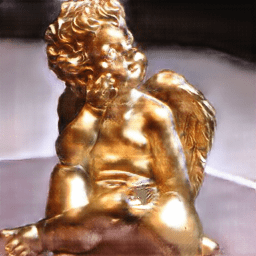} &
\includegraphics[width=0.19\linewidth]{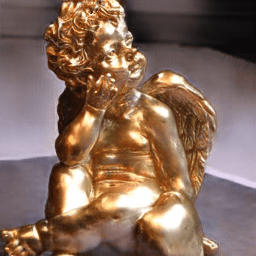} &
\includegraphics[width=0.19\linewidth]{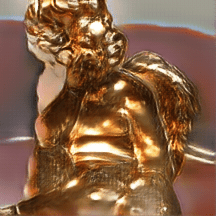} &
\includegraphics[width=0.19\linewidth]{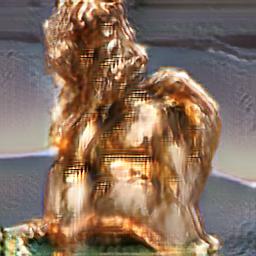} \\

& \FID=\textbf{198.91} & \FID=223.23 & \FID=218.11 & \FID=228.99\\

\end{tabular}
\caption{Specular $\leftrightarrow$ Diffuse translation: Columns show the inputs and various competing results. The two top rows show a translation from a specular input to a diffuse output, while the two bottom rows show the opposite translation.}
\label{tab:Spec2Diff}
\end{figure}

In Figure \ref{tab:Spec2Diff} we show the results of applying \OurName on Wu et al.'s dataset.
Results are shown for both translation directions, specular to diffuse and diffuse to specular. Both types of translations produce visually convincing results. It may be seen that our translation captures the fine details of the sculptures, and produces proper shading. An unintended phenomenon, are the changes which are created in the background. A perfect translation between these two domains shouldn't have altered any background pixels, but our method provides no means of controlling which pixels are left untouched.
Additional results are provided in the supplementary material.

\subsection{Mobile phone to SLR}
Although extremely popular, contemporary mobile phones are still very far from being able to produce results of quality comparable to those of a professional DSLR camera. This is mostly due to the limitations on sensor and aperture size. In a recent work, Ignatov et al.~\cite{ignatov2017dslr} propose a supervised method, and apply it on their own manually collected large-scale dataset, consisting of real photos captured by three different phones and one high-end reflex camera. Here, we demonstrate our approach's applicability on the same task.


Ignatov et al.'s approach is fully supervised, where pairing of images (mobile and DSLR photos) is available during training. To achieve this, they introduced a large-scale DPED dataset that consists of photos taken synchronously in the wild by a smartphone and a DSLR camera. In contrast to DPED, we naively train \OurName to translate iPhone photos to those of DSLR quality, ignoring the pairing and adding no additional explicit priors on the data.

Since the DLSR camera did not capture the scene from exactly the same perspective as the iPhone, a quantitative comparison is not possible. Ignatov at al. proposed aligning and warping between the two images, but since the entire essence of the approaches we compare is to produce high resolution details, we find warping (and thus resampling) improper and resort to a qualitative comparison. However, we report the FID as an objective method of comparison.

We compare our results to CycleGAN, MUNIT \cite{huang2018multimodal}, DRIT \cite{lee2018diverse} and to the fully supervised DPED method, and show that our approach generates translations that are richer in colors and details. The FID results show that DPED's supervised method yields the best results, but that among the unsupervised methods, our results ranks first.

\begin{figure}[ht]
\begin{tabular}{cccccc}

Input & \OurName & CycGAN & MUNIT & DRIT & DPED \\

\includegraphics[width=0.16\linewidth]{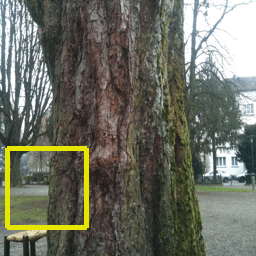} &
\includegraphics[width=0.16\linewidth]{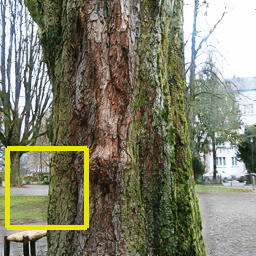} &
\includegraphics[width=0.16\linewidth]{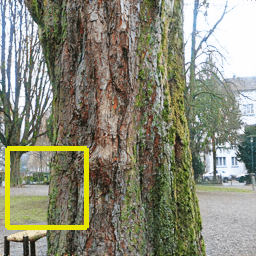} &
\includegraphics[width=0.16\linewidth]{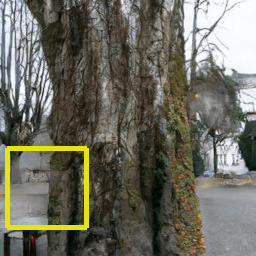} & 
\includegraphics[width=0.16\linewidth]{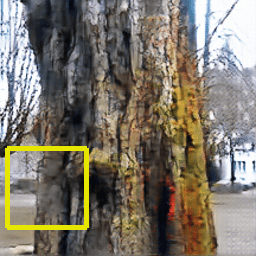} &
\includegraphics[width=0.16\linewidth]{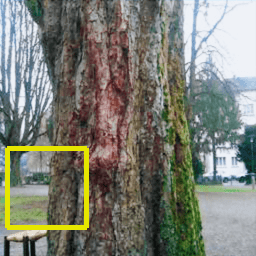} \\

\includegraphics[width=0.16\linewidth]{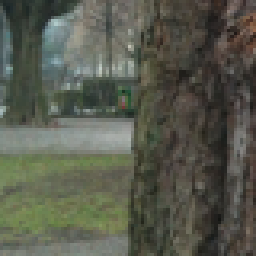} &
\includegraphics[width=0.16\linewidth]{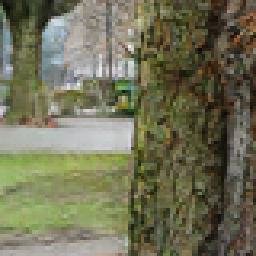} &
\includegraphics[width=0.16\linewidth]{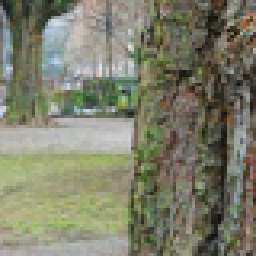} &
\includegraphics[width=0.16\linewidth]{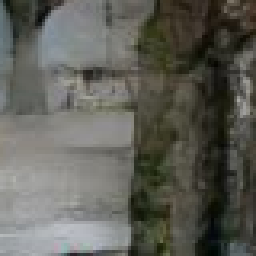} &
\includegraphics[width=0.16\linewidth]{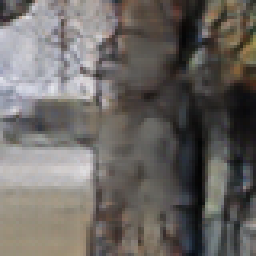} & 
\includegraphics[width=0.16\linewidth]{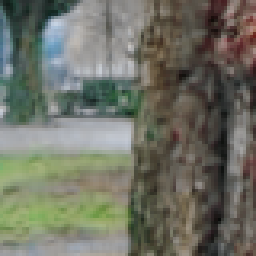} \\

\includegraphics[width=0.16\linewidth]{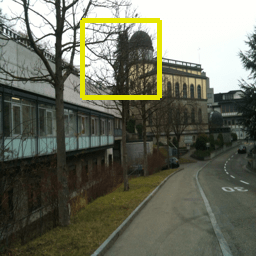} &
\includegraphics[width=0.16\linewidth]{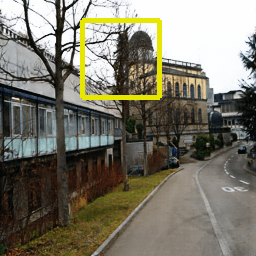} &
\includegraphics[width=0.16\linewidth]{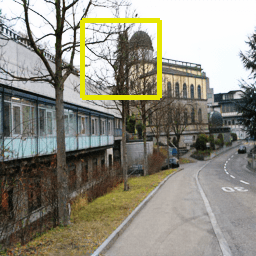} &
\includegraphics[width=0.16\linewidth]{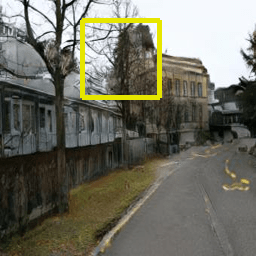} & 
\includegraphics[width=0.16\linewidth]{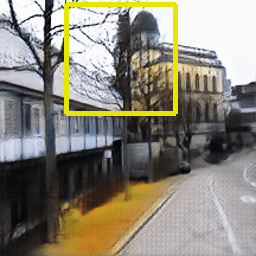} &
\includegraphics[width=0.16\linewidth]{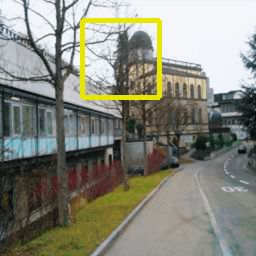} \\

\includegraphics[width=0.16\linewidth]{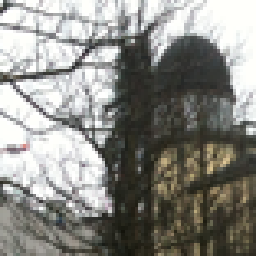} &
\includegraphics[width=0.16\linewidth]{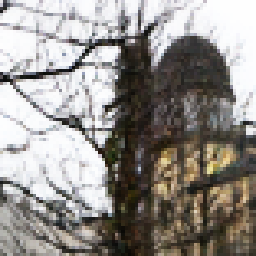} &
\includegraphics[width=0.16\linewidth]{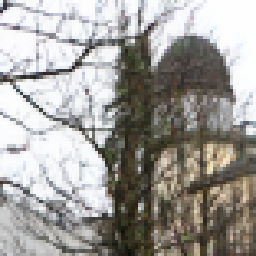} &
\includegraphics[width=0.16\linewidth]{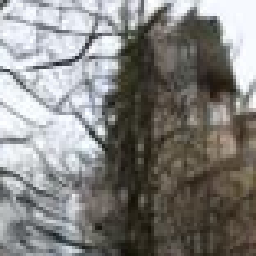} &
\includegraphics[width=0.16\linewidth]{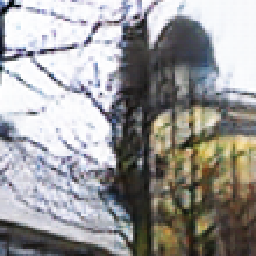} &
\includegraphics[width=0.16\linewidth]{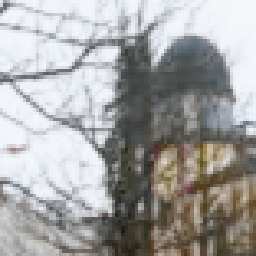} \\

\includegraphics[width=0.16\linewidth]{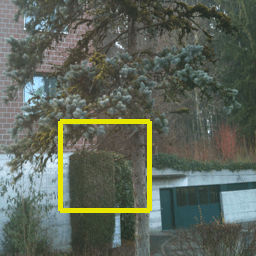} &
\includegraphics[width=0.16\linewidth]{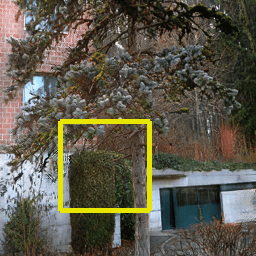} &
\includegraphics[width=0.16\linewidth]{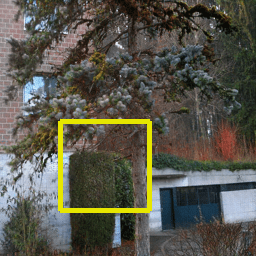} &
\includegraphics[width=0.16\linewidth]{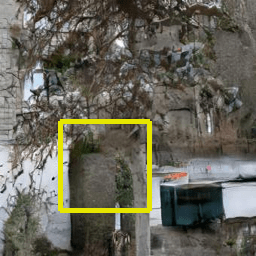} &
\includegraphics[width=0.16\linewidth]{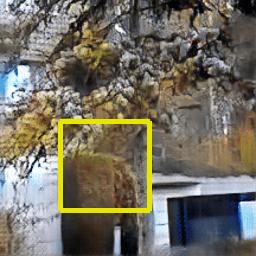} &
\includegraphics[width=0.16\linewidth]{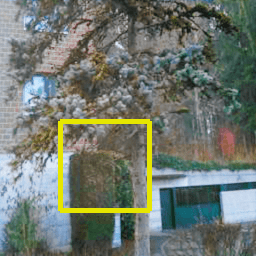} \\

\includegraphics[width=0.16\linewidth]{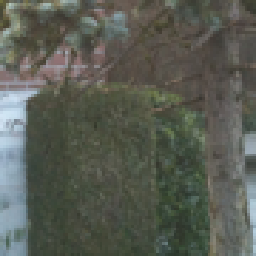} &
\includegraphics[width=0.16\linewidth]{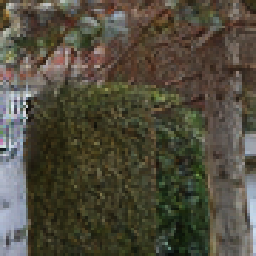} &
\includegraphics[width=0.16\linewidth]{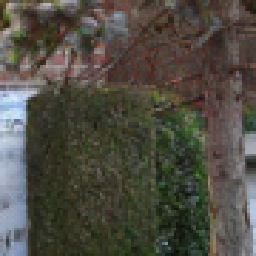} &
\includegraphics[width=0.16\linewidth]{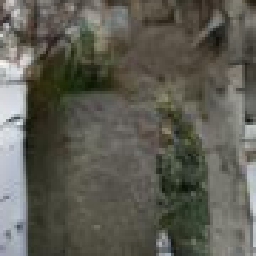} &
\includegraphics[width=0.16\linewidth]{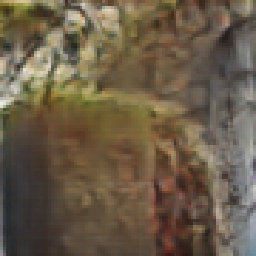} &
\includegraphics[width=0.16\linewidth]{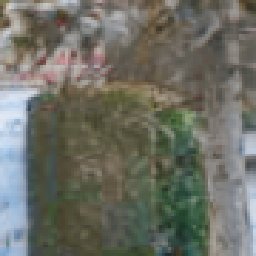} \\

& \FID=161.81 & \FID=181.68 & \FID=182.21 & \FID=163.41 & \BFFID=\textbf{81.12}\\

\end{tabular}
\caption{Results of mobile phone photo enhancement. We show the input, \OurName, CycleGAN's, MUNIT's, DRIT's and DPED's results, from left to right. Odd rows show full images, while even ones show a zoom-in.}
\label{tab:Mob2SLRZoom}
\vspace{-4mm}
\end{figure}

In Figure \ref{tab:Mob2SLRZoom} we present our results and compare them with those mentioned above. Zooming into specific regions of the results, shows the effectiveness of \OurName vs. the competitors, for producing colors and details in the translated images. In order to properly compare, all of the algorithms were fed with an input image scaled to 256x256.
We show the input, \OurName, CycleGAN's, MUNIT's, DRIT's and DPED's results, from left to right respectively. The odd rows show the full images, while the even rows show a zoom-in. The two top rows show \OurName's ability to enhance colors better than the competing methods. The two middle rows show that our approach enhances contrast more strongly. Finally, the two bottom rows show that our result enhances details better than the competitors.
We provide additional results in the supplementary material.

\setlength\tabcolsep{0.5pt}

%% file: appl_seg.tex
\subsection{Semantic foreground extraction}


Here, we demonstrate that using \OurName it is possible to learn to extract a common object of interest from an image, by training on an unstructured image collection obtained using internet search engines.
Specifically, we search Google images, once by using the search term ``elephant'', and once by using ``elephant white background'' to obtain two sets of images, with roughly 500 images in each set. \OurName is then trained to translate between these two sets in an unpaired fashion. Note that this is a highly asymmetric translation task: while replacing all non-elephant pixels in an image with white color is well-defined, the opposite direction has a multitude of plausible outcomes. Thus, one can hardly expect a bijection between the original image domains, underscoring the importance of constraining and regularizing the latent spaces.

To obtain clearly extracted objects, we employ a slightly modified pipeline. Rather than using the raw outcome of the translation from images with a general background to ones with a white background, we use the result to define a segmentation mask by simple thresholding. Specifically, we replace near-white pixels with white color, while the color of the remaining pixels is replaced with their original values (before the translation). A pixel is considered nearly white if its luminance exceeds 243.

\begin{figure}[ht]
\begin{tabular}{cccccc}

\footnotesize{Input} & \footnotesize{\OurName} & \footnotesize{CycleGAN} & \footnotesize{GrabCut} & \footnotesize{AGGAN} & \footnotesize{AGGAN Mask}\\

\includegraphics[width=0.16\linewidth]{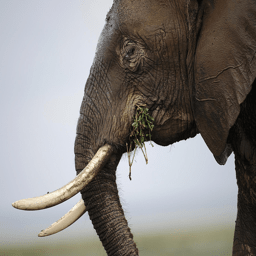} &
\includegraphics[width=0.16\linewidth]{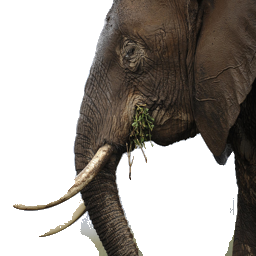} &
\includegraphics[width=0.16\linewidth]{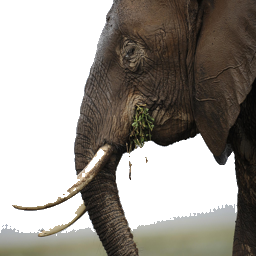} & 
\includegraphics[width=0.16\linewidth]{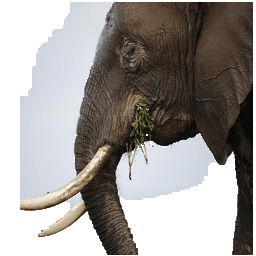} & 
\includegraphics[width=0.16\linewidth]{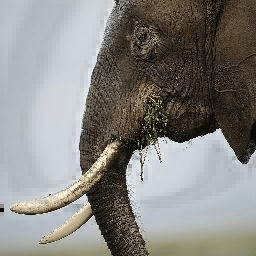} & 
\includegraphics[width=0.16\linewidth]{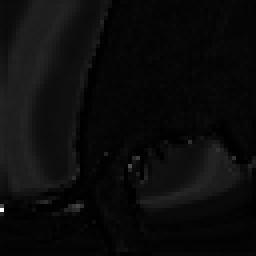} \\

\includegraphics[width=0.16\linewidth]{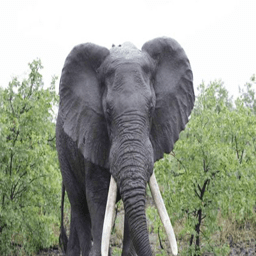} &
\includegraphics[width=0.16\linewidth]{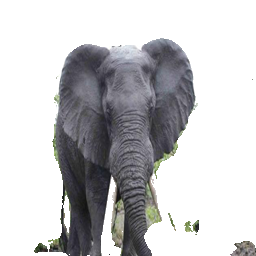} &
\includegraphics[width=0.16\linewidth]{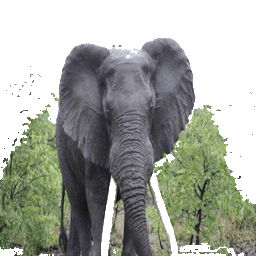} & 
\includegraphics[width=0.16\linewidth]{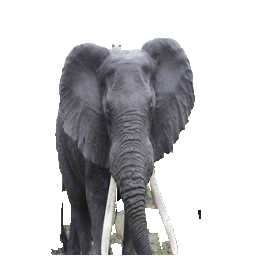} & 
\includegraphics[width=0.16\linewidth]{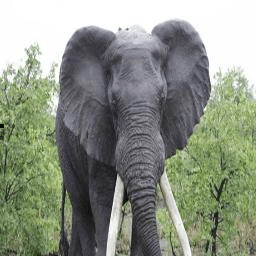} & 
\includegraphics[width=0.16\linewidth]{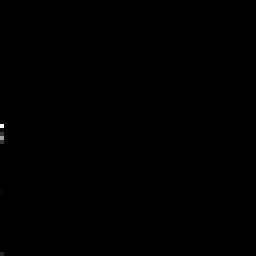} \\

\includegraphics[width=0.16\linewidth]{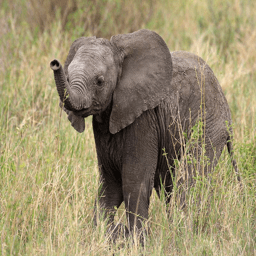} &
\includegraphics[width=0.16\linewidth]{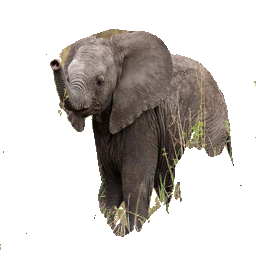} &
\includegraphics[width=0.16\linewidth]{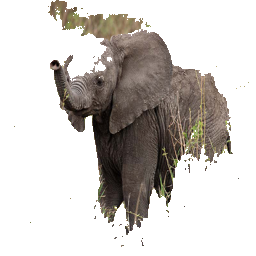} & 
\includegraphics[width=0.16\linewidth]{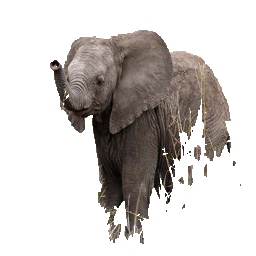} & 
\includegraphics[width=0.16\linewidth]{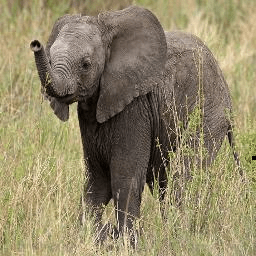} & 
\includegraphics[width=0.16\linewidth]{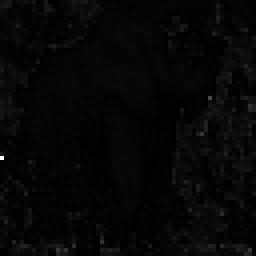} \\

\end{tabular}
\caption{Results of employing \OurName for foreground extraction, applied on elephant photos. From left to right, we show the input image, CrossNet, CycleGAN, GrabCut and AGGAN.}
\label{tab:Eleph}
\vspace{-2mm}
\end{figure}

Figure \ref{tab:Eleph} presents some of our results, applied on test images of elephants. We compare our results with CycleGAN, which were applied with the same post processing operation as our results. We also compare with AGGAN~\cite{mejjati2018unsupervised}, which extracts attention maps which have been shown to be useful for segmentation. Additionally, we also compare our method with GrabCut \cite{rother2004grabcut}. Since GrabCut requires a initial bounding box, we provide it manually for the test images.

The results in Figure \ref{tab:Eleph} show that although our method assumes no priors on the segmentation task at hand, it effectively extracts the foreground objects. We attribute this success to the presence of all of the necessary information in the latent code, making our approach more adequate for this task. AGGAN is shown to fail on this task, most probably due to the same class of images found in the two input sets, confusing its attention mechanism. Note how its attention masks do not indicate any clear attention region. Additional results are provided in the supplementary material.

In addition to our manually composed dataset of elephants, we quantify our method's results by employing the UT Zappos50K shoes dataset \cite{finegrained} which present photos with white backgrounds as well as on the HDSeg dataset \cite{kolesnikov2014closed} of Horses with ground truth segmentation masks. During training to segment shoes, we synthetically replaced the white background with natural scenery photos.

\begin{figure}
\centering
\begin{tabular}{cc}
\includegraphics[width=0.475\linewidth, height=95px]{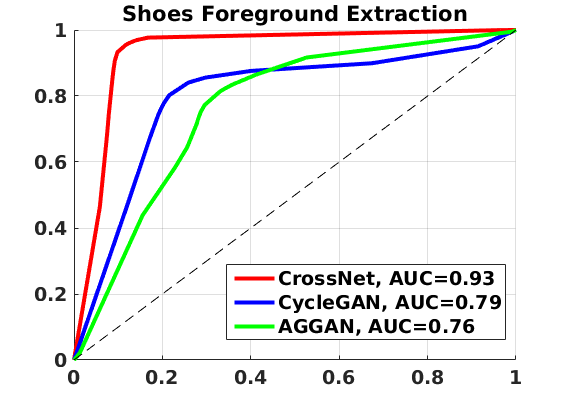} & 
\includegraphics[width=0.475\linewidth, height=95px]{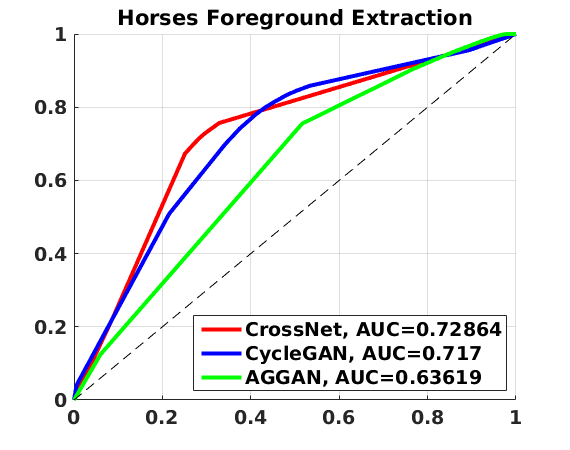} \\
\end{tabular}
\caption{Objective evaluations comparing between \OurName, CycleGAN and AGGAN for foreground extraction on the UT Zappos50K dataset and HDSeg.}
\label{fig:SegPerc}
\end{figure}

We report the RoC curves of \OurName, CycleGAN and AGGAN in Figure \ref{fig:SegPerc}, as well as the Area Under Curve (AUC) through which \OurName's superiority is noticeable.

%% file: conclusions.tex
\section{Discussion and Future work}

We have presented \OurName, an architecture for generic unpaired image-to-image translation. During training, \OurName imposes constraints on the latent codes produced by the generators, rather than imposing them in the image domains. Latent codes constitute a compact representation of the aspects of the data that are most relevant to the task that the model is trained for. Since the unpaired translation setting is challenging and under-constrained, proper regularization of these latent spaces is crucial, and we demonstrate that models trained with such regularizers are more effective at their intended task. 


As a generic method, naturally, no prior on the specific task is used. We believe that through the addition of task specific priors, one may leverage CrossNet and push the visual quality of the results further, for specific tasks. We see this path as a promising topic for further research.

An interesting observation is that all our losses are symmetric. They operate in both directions, $A$ to $B$ and $B$ to $A$. Although we have shown pleasing results when applying our method to the proposed applications, in some of them, the asymmetric nature of the task suggests that asymmetric constraints might be more appropriate.
For example, in the foreground extraction application, background removal is well defined, but adding a background is not, having a huge space of plausible outcomes. The exploration of asymmetric architectures and constraints is another promising direction for future work.

%
%
%
%
%

%% file: main.bbl
\begin{thebibliography}{10}\itemsep=-1pt

\bibitem{choi2017stargan}
Y.~Choi, M.~Choi, M.~Kim, J.-W. Ha, S.~Kim, and J.~Choo.
\newblock Stargan: Unified generative adversarial networks for multi-domain
  image-to-image translation.
\newblock {\em arXiv preprint}, 1711, 2017.

\bibitem{dong2017unsupervised}
H.~Dong, P.~Neekhara, C.~Wu, and Y.~Guo.
\newblock Unsupervised image-to-image translation with generative adversarial
  networks.
\newblock {\em arXiv preprint arXiv:1701.02676}, 2017.

\bibitem{goodfellow2014generative}
I.~Goodfellow, J.~Pouget-Abadie, M.~Mirza, B.~Xu, D.~Warde-Farley, S.~Ozair,
  A.~Courville, and Y.~Bengio.
\newblock Generative adversarial nets.
\newblock In {\em Advances in neural information processing systems}, pages
  2672--2680, 2014.

\bibitem{heusel2017gans}
M.~Heusel, H.~Ramsauer, T.~Unterthiner, B.~Nessler, and S.~Hochreiter.
\newblock Gans trained by a two time-scale update rule converge to a local nash
  equilibrium.
\newblock In {\em Advances in Neural Information Processing Systems}, pages
  6626--6637, 2017.

\bibitem{hoshen2018nam}
Y.~Hoshen and L.~Wolf.
\newblock Nam: Non-adversarial unsupervised domain mapping.
\newblock {\em arXiv preprint arXiv:1806.00804}, 2018.

\bibitem{huang2018multimodal}
X.~Huang, M.-Y. Liu, S.~Belongie, and J.~Kautz.
\newblock Multimodal unsupervised image-to-image translation.
\newblock {\em arXiv preprint arXiv:1804.04732}, 2018.

\bibitem{ignatov2017dslr}
A.~Ignatov, N.~Kobyshev, R.~Timofte, K.~Vanhoey, and L.~Van~Gool.
\newblock Dslr-quality photos on mobile devices with deep convolutional
  networks.
\newblock In {\em the IEEE Int. Conf. on Computer Vision (ICCV)}, 2017.

\bibitem{isola2017image}
P.~Isola, J.-Y. Zhu, T.~Zhou, and A.~A. Efros.
\newblock Image-to-image translation with conditional adversarial networks.
\newblock {\em arXiv preprint}, 2017.

\bibitem{johnson2016perceptual}
J.~Johnson, A.~Alahi, and L.~Fei-Fei.
\newblock Perceptual losses for real-time style transfer and super-resolution.
\newblock In {\em European Conference on Computer Vision}, pages 694--711.
  Springer, 2016.

\bibitem{kim2017learning}
T.~Kim, M.~Cha, H.~Kim, J.~K. Lee, and J.~Kim.
\newblock Learning to discover cross-domain relations with generative
  adversarial networks.
\newblock {\em arXiv preprint arXiv:1703.05192}, 2017.

\bibitem{kolesnikov2014closed}
A.~Kolesnikov, M.~Guillaumin, V.~Ferrari, and C.~H. Lampert.
\newblock Closed-form approximate crf training for scalable image segmentation.
\newblock In {\em European Conference on Computer Vision}, pages 550--565.
  Springer, 2014.

\bibitem{lee2018diverse}
H.-Y. Lee, H.-Y. Tseng, J.-B. Huang, M.~Singh, and M.-H. Yang.
\newblock Diverse image-to-image translation via disentangled representations.
\newblock In {\em Proceedings of the European Conference on Computer Vision
  (ECCV)}, pages 35--51, 2018.

\bibitem{li2016precomputed}
C.~Li and M.~Wand.
\newblock Precomputed real-time texture synthesis with markovian generative
  adversarial networks.
\newblock In {\em European Conference on Computer Vision}, pages 702--716.
  Springer, 2016.

\bibitem{liu2017unsupervised}
M.-Y. Liu, T.~Breuel, and J.~Kautz.
\newblock Unsupervised image-to-image translation networks.
\newblock In {\em Advances in Neural Information Processing Systems}, pages
  700--708, 2017.

\bibitem{liu2016coupled}
M.-Y. Liu and O.~Tuzel.
\newblock Coupled generative adversarial networks.
\newblock In {\em Advances in neural information processing systems}, pages
  469--477, 2016.

\bibitem{mao2017least}
X.~Mao, Q.~Li, H.~Xie, R.~Y. Lau, Z.~Wang, and S.~P. Smolley.
\newblock Least squares generative adversarial networks.
\newblock In {\em Computer Vision (ICCV), 2017 IEEE International Conference
  on}, pages 2813--2821. IEEE, 2017.

\bibitem{mejjati2018unsupervised}
Y.~A. Mejjati, C.~Richardt, J.~Tompkin, D.~Cosker, and K.~I. Kim.
\newblock Unsupervised attention-guided image-to-image translation.
\newblock In {\em Advances in Neural Information Processing Systems}, pages
  3697--3707, 2018.

\bibitem{rother2004grabcut}
C.~Rother, V.~Kolmogorov, and A.~Blake.
\newblock Grabcut: Interactive foreground extraction using iterated graph cuts.
\newblock In {\em ACM transactions on graphics (TOG)}, volume~23, pages
  309--314. ACM, 2004.

\bibitem{shrivastava2017learning}
A.~Shrivastava, T.~Pfister, O.~Tuzel, J.~Susskind, W.~Wang, and R.~Webb.
\newblock Learning from simulated and unsupervised images through adversarial
  training.
\newblock In {\em CVPR}, volume~2, page~5, 2017.

\bibitem{srivastava2017veegan}
A.~Srivastava, L.~Valkov, C.~Russell, M.~U. Gutmann, and C.~Sutton.
\newblock Veegan: Reducing mode collapse in gans using implicit variational
  learning.
\newblock In {\em Advances in Neural Information Processing Systems}, pages
  3308--3318, 2017.

\bibitem{taigman2016unsupervised}
Y.~Taigman, A.~Polyak, and L.~Wolf.
\newblock Unsupervised cross-domain image generation.
\newblock 2017.

\bibitem{wu2018specular}
S.~Wu, H.~Huang, T.~Portenier, M.~Sela, D.~Cohen-Or, R.~Kimmel, and M.~Zwicker.
\newblock Specular-to-diffuse translation for multi-view reconstruction.
\newblock In {\em European Conference on Computer Vision}, pages 193--211.
  Springer, 2018.

\bibitem{yi2017dualgan}
Z.~Yi, H.~R. Zhang, P.~Tan, and M.~Gong.
\newblock Dualgan: Unsupervised dual learning for image-to-image translation.
\newblock In {\em ICCV}, pages 2868--2876, 2017.

\bibitem{finegrained}
A.~Yu and K.~Grauman.
\newblock Fine-grained visual comparisons with local learning.
\newblock In {\em Computer Vision and Pattern Recognition (CVPR)}, Jun 2014.

\bibitem{zhu2017unpaired}
J.-Y. Zhu, T.~Park, P.~Isola, and A.~A. Efros.
\newblock Unpaired image-to-image translation using cycle-consistent
  adversarial networks.
\newblock {\em arXiv preprint}, 2017.

\end{thebibliography}
